%% file: iros2016.tex
\documentclass[letterpaper, 10 pt, conference]{ieeeconf}  

\usepackage{times}
\usepackage{textcomp}
\usepackage{amsmath}
\usepackage{amssymb}
\usepackage{graphicx}
\usepackage[pagebackref=true,breaklinks=true,letterpaper=true,colorlinks,bookmarks=false]{hyperref}
\usepackage{booktabs}     
\usepackage{enumerate}    
\usepackage{cite}         
\usepackage{tikz}
\usepackage{fix2col}
\usepackage{soul}         

\usetikzlibrary{arrows,positioning} 
\usetikzlibrary{calc,patterns,decorations.pathmorphing,decorations.markings,fit,backgrounds}
\tikzset{
    >=stealth',
    box/.style={
           rectangle,
           rounded corners,
           draw=black, very thick,
                          minimum height=1cm,
                          minimum width=1.5cm,
           text centered},
    state/.style={
                rectangle,
                rounded corners,
                draw=black, very thick,
                minimum height=1.0cm,
                minimum width=1.0cm,
                text centered},
    pil/.style={
           ->,
           thick,
           shorten <=2pt,
           shorten >=6pt,}

}

\IEEEoverridecommandlockouts                              

\title{\LARGE \bf Multi-Lane Perception Using Feature Fusion Based on GraphSLAM}

\author{Alexey Abramov$^{\S}$, Christopher Bayer$^{\S}$, Claudio Heller$^{\S}$, Claudia Loy$^{\S}$
\thanks{${\S}$ These authors contributed equally to this work.}
\thanks{Alexey Abramov, Christopher Bayer, Claudio Heller, Claudia Loy are with Continental Teves AG, Chassis \& Safety Division, Advanced Engineering, Guerickestrasse 7, DE-60488, Frankfurt am Main, Germany. \tt\small{\{alexey.abramov,christopher.bayer, claudio.heller,claudia.loy\}@continental- corporation.com}}
}

\begin{document}

\maketitle
\thispagestyle{empty}
\pagestyle{empty}

\begin{abstract}
An extensive, precise and robust recognition and modeling of the environment is a key factor for next generations of Advanced Driver Assistance Systems and development of autonomous vehicles. In this paper, a real-time approach for the perception of multiple lanes on highways is proposed. Lane markings detected by camera systems and observations of other traffic participants provide the input data for the algorithm. The information is accumulated and fused using GraphSLAM and the result constitutes the basis for a multi-lane clothoid model. To allow incorporation of additional information sources, input data is processed in a generic format. Evaluation of the method is performed by comparing real data, collected with an experimental vehicle on highways, to a ground truth map. The results show that ego and adjacent lanes are robustly detected with high quality up to a distance of 120~m. In comparison to serial lane detection, an increase in the detection range of the ego lane and a continuous perception of neighboring lanes is achieved. The method can potentially be utilized for the longitudinal and lateral control of self-driving vehicles.
\end{abstract}

\section{INTRODUCTION AND RELATED WORK \label{introduction}}
\input{Introduction.tex}

\section{FEATURE BASED LANE ESTIMATION \label{methods}}
\input{LaneEstimationMethod.tex}

\subsection{Experimental vehicle and sensor setup \label{experimental-setup}}
\input{Experimentalsetup.tex}

\subsection{Camera-based lane detection and feature extraction \label{camera-based-lane-detection}} 
\input{Camerafeatures.tex}

\subsection{Traffic participant detection \label{data-object-fusion}} 
\input{Objectfusion.tex}

\subsection{Lane feature fusion using GraphSLAM \label{graph-slam}} 
\input{Graphslam.tex}

\subsection{Building the graph \label{building-graph}}
\input{Graphbuilding.tex}

\subsection{Multi-lane modeling \label{road-model}} 
\input{Roadmodel.tex}

\section{EXPERIMENTAL EVALUATION \label{evaluation}}


\input{Evaluatiomethod.tex}



\section{CONCLUSION \label{conclusion}}
\input{Conclusion.tex}




{\small
\bibliographystyle{IEEEtran}
\bibliography{iros2016_references}
}

\end{document}

%% file: Introduction.tex
Nowadays, modern cars are equipped with Advanced Driver Assistance Systems (ADAS) to increase the comfort and safety of drivers and passengers. Systems like lane departure warning, lane keeping assist, adaptive cruise control, emergency brake assist and blind spot monitoring help drivers to keep the car within the lane and avoid collisions with other traffic participants~\cite{Bengler2014}. For the next generation of ADAS, and especially regarding the prospect of autonomous vehicles, comprehensive and extensive knowledge about the environment is required. One fundamental part is a continuous and robust perception of the road. This includes extension of both longitudinal and lateral detection range with perception of all traffic lanes relevant for analysis of the current driving situation. 

For sensing an ego lane state-of-the-art ADAS rely on camera systems to detect left and right markings~\cite{Dickmanns1992,Wang2000,McCall2006,Meis2010}. Despite recent developments in vision-based lane detection techniques~\cite{Aly2008,Kim2008,Liu2011,Kang2014} the performance of camera systems remains limited to the camera's aperture angle, variations in illumination, quality of lane markings and weather conditions. To detect other traffic participants serial vehicles utilize camera, radar, ultrasonic and lidar sensors. 

Recent research demonstrates that fusion of measurements from diverse sensors ensures a more sophisticated representation of the environment~\cite{Leonard2008,Fatemi2014,Cho2014}. This involves precise and robust lane estimation which is essential for automated driving at high speeds and automated lane changes. Common multi-sensor fusion techniques combine camera sensors with measurements of radar, lidar or both in order to compensate drawbacks of each other~\cite{Huang2009,Fatemi2014}.

Existing real-time multi-lane detection systems rely on optical sensors. \textit{Aly et~al.}~\cite{Aly2008} introduced a camera-based approach which detects and models multiple lanes in still images. However, this technique works well only on free roads as it does not take other traffic participants into account~\footnote{Passing vehicles can cause diagonal lines in the Inverse Perspective Mapping (IPM) that eventually lead to wrong lane markings.}. In addition, lines with prominent contrast oriented parallel to the road direction (e.g. street writing, shadows of guardrails, curbs, etc.) can lead to false detections because the real width of lane markings is not considered~\cite{Aly2008Online}. A camera-based system proposed by \textit{Kang et~al.}~\cite{Kang2014} detects up to six lanes on highways. Nevertheless, this method also neglects information about other traffic participants and, thereby, lanes covered by other cars cannot be detected. Furthermore, it employs a second order polynomial with constant curvature to model lane markings. \textit{Huang et~al.}~\cite{Huang2009} detect multiple lanes using camera and lidar data, where lidar is employed as a complementary sensor for eliminating outliers~\footnote{Objects that can generate wrong lane detections are detected by the lidar-based obstacle detection. All 3D obstacle detections are projected onto the camera image to suppress explicitly road line candidates detected in those areas.}. A high detection performance is achieved with a setup of 5 cameras and 13 lidars. Nevertheless, typical problems of optical sensors, such as perception of hidden or poorly visible lane markings, remain unresolved.

In this paper, a novel real-time multi-lane perception approach for highways is presented, which provides a continuous and robust estimation of ego and adjacent lanes up to $120~m$. The main advantage of the proposed technique is detection of multiple lanes combining data from various input sources. For this purpose a generic feature description is introduced which allows processing and fusion of measurements regardless of the sensor type. In the present study, lane markings detected by two camera systems are fused with tracked trajectories of other traffic participants. The fusion result is utilized for estimation of a multi-lane clothoid model, which can serve as an input for the longitudinal and lateral control of self-driving vehicles. The system performance is evaluated by comparing real data collected on highways to ground truth. This multi-lane perception approach operates without prior knowledge about course of the road and number of existing traffic lanes. One more important advantage over existing methods is that the demonstrated application relies on a production-oriented sensor setup and requires neither digital map nor GPS localization~\cite{Brubaker2015}.

A description of the lane estimation algorithm can be found in section~\ref{methods} and the evaluation methods and results are presented in section \ref{evaluation} followed by the conclusion in section \ref{conclusion}.

%% file: LaneEstimationMethod.tex
\begin{figure}
\begin{tikzpicture}[->,>=stealth',shorten >=1pt,auto,node distance=3cm,
  thick]

  \node[state] (S1) [] {$S_1$};
  \node[state] (S2) [right of=S1,  node distance=1.5 cm]{$S_2$};
  \node[state] (S3) [right of=S2,  node distance=1.5 cm]{$S_3$};
  \node[] (SDots) [right of=S3,  node distance=1.0 cm]{$...$};
  \node[state] (SN) [right of=SDots, node distance=1.0cm] {$S_N$};

  \node[state] (FF) [below right=1cm and 0.6cm of S1] {lane feature fusion};
  \node[state] (RM) [below of=FF,  node distance=2.0 cm]{multi-lane modeling};

  \path[every node/.style={font=\sffamily\small}]
    (S1) edge[black] node [left] {$\mathbf{F}_1$} (FF)
    (S2) edge[black] node [left] {$\mathbf{F}_2$} (FF)
    (S3) edge[black] node [left] {$\mathbf{F}_3$} (FF)
    (SN) edge[black] node [left] {$\mathbf{F}_N$} (FF)
    (FF) edge[black] node [left] {$\mathbf{F_{fused}}$} (RM)
    ;

    
\end{tikzpicture}
\centering
\caption{Architectural overview of the multi-lane perception method. $\mathbf{F_1},... \mathbf{F_N}$ are feature sets obtained from different input sources $S_1,...,S_N$, $\mathbf{F_{fused}}$ represents a set of fused lane features.
}

\label{fig:architecture}
\end{figure}
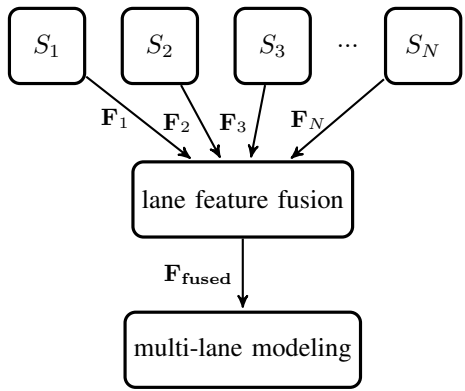
The principal ideas of the multi-lane perception methods are described in the following. An overview of the main modules is shown in fig.~\ref{fig:architecture}, where $S_1,...,S_N$ represent $N$ different input sources, which allow deriving information about the road course (see sections~\ref{camera-based-lane-detection} and~\ref{data-object-fusion}). The core module is the lane feature fusion algorithm that accumulates and fuses the feature sets $\mathbf{F_1},... \mathbf{F_N}$ obtained from the input sources. This method combines advantages of every input data type and is explained in detail in sections~\ref{graph-slam} and~\ref{building-graph}. On top of the fusion result, a clothoid based multi-lane model representing the road course is estimated (see section~\ref{road-model}).

Hence the lane feature fusion is free of geometric model assumptions. To ensure a generic representation in the lane feature fusion a uniform interface is utilized for the inputs $\mathbf{F_1},... \mathbf{F_N}$ and the output $\mathbf{F_{fused}}$. The interface is specified as a list of features
\begin{equation}
	\mathbf{f}_i = [x_i,y_i,\theta_i,c_i,\mathbf{\Sigma}_i],
	\label{def.feature}
\end{equation}
where $x_i$ and $y_i$ constitute the position of a feature $\mathbf{f}_{i}$ and $\theta_i$ constitutes its heading. The measurement uncertainty of each feature is given by a confidence value $c_i \in [0,1]$ and a covariance matrix $\mathbf{\Sigma}_i \in \mathbb{R}^{3\times3}$ with respect to $x$, $y$ and $\theta$. Note that this is a two-dimensional representation which omits height information about the road course.

%% file: Experimentalsetup.tex
The experimental vehicle used in the current study is equipped with the following sensors for environmental perception:

\begin{itemize}
	\item Serial Mono Camera (SMC) mounted behind the windshield above the rear-view mirror~\footnote{Here the right camera of Stereo Multi Functional Camera (SMFC) 300 series by Continental is used.}.
	\item High Resolution Camera (HRC) mounted close to the SMC behind the windshield~\footnote{USB 3.0 camera UI-3580CP, IDS Imaging Development Systems GmbH.}.
	\item Long Range Radar (LRR) mounted inside the front bumper below the license plate~\footnote{ARS 300, Continental, range $200~m$.}.
	\item Short Range Radars (SRR) mounted inside the front and rear bumper at each corner~\footnote{SRR, Continental, range $50~m$.}
\end{itemize}

The vehicle coordinate system $(x,y,z)$ is a right-handed coordinate system, its origin lies in the middle of the front axle (height of the road), $x$ is identical to the driving direction, $y$ points to the left and $z$ points upwards.



%% file: Camerafeatures.tex
\begin{figure}[!ht]
	\begin{center}
		\includegraphics[width=8.6cm]{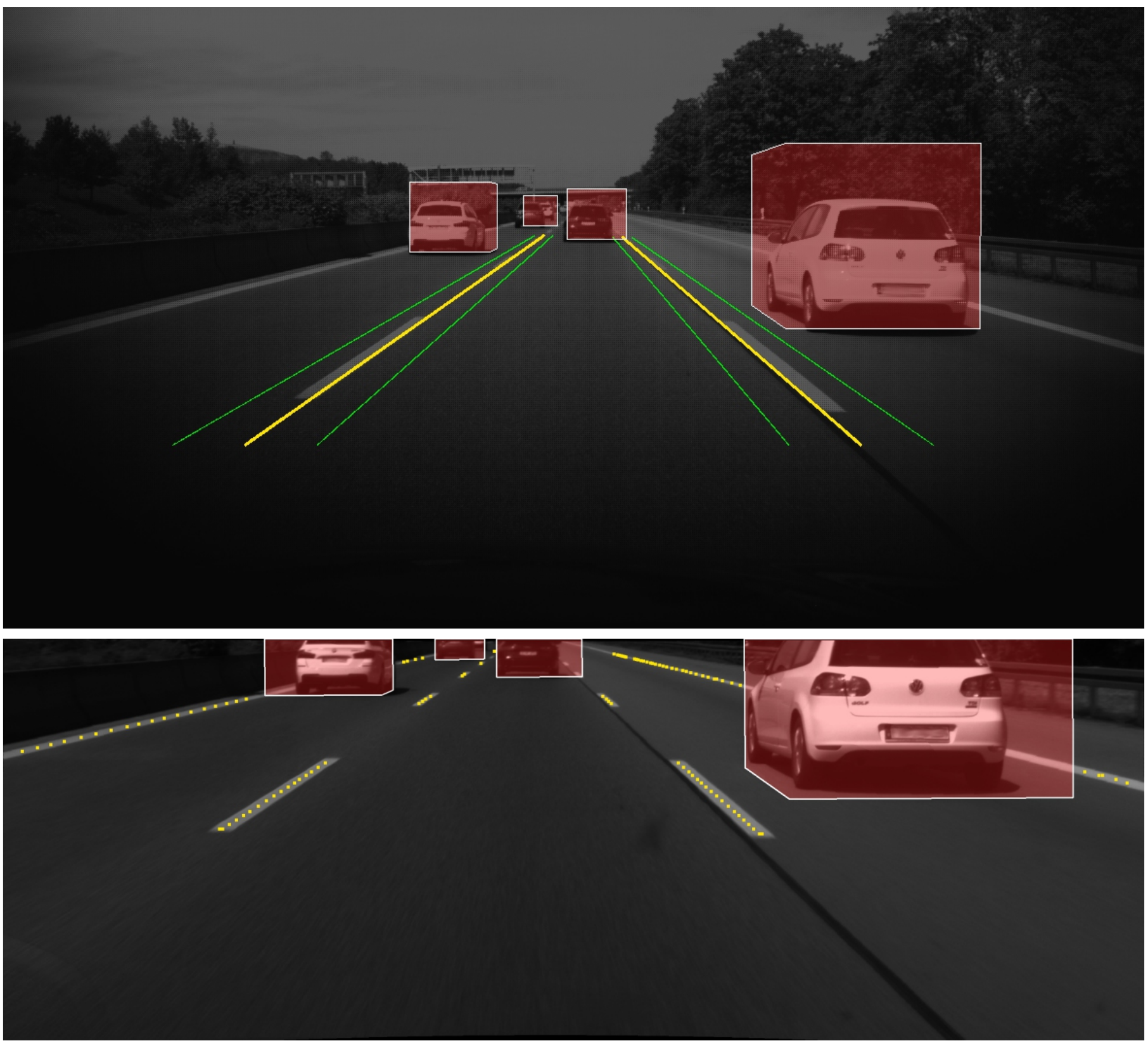}
		\caption{Perception of lane markings and moving objects (red boxes) by the experimental vehicle. The ego lane estimated by the SMC is shown in the top (lane clothoids in yellow, adoptive measurement windows in green), while lane marking features detected by the HRC are shown at the bottom (yellow points).
			\label{fig:lane-detection}}
	\end{center}
\end{figure}

The SMC is a serial automotive camera with a resolution of $1176 \times 640$ px and an aperture angle of $53$\textdegree. It performs multiple functions such as lane detection, traffic sign and pedestrian recognition and detection of other vehicles. The SMC lane detection is a combination of several image processing operations followed by a Kalman filter used for tracking~\cite{Meis2010}. The inner side of left and right lane markings of the ego lane are represented by clothoids (see fig.~\ref{fig:lane-detection} in the top). The algorithm has a detection range up to $90~m$ under ideal conditions (good visible markings and appropriate weather and lighting).

The HRC is an experimental camera with a resolution of $2560 \times 900$ px and an aperture angle of $48$\textdegree. In this work it is used to increase the detection range of the ego lane and, in addition, to extract lane markings from neighboring lanes. The algorithm processing the HRC images is based on the serial SMC lane detection and finds features at lane markings taking their physical sizes into account (shown in fig.~\ref{fig:lane-detection} at the bottom). Note that no feature tracking or lane modeling are performed based on the HRC images alone. Detected features directly constitute the input for the lane feature fusion. Under good conditions the HRC features can be generated up to a longitudinal distance of $130~m$.

Both cameras use online calibration algorithms to transform the SMC ego lane and marking features from the HRC to the vehicle coordinate system~\cite{Liang2004}.

%% file: Objectfusion.tex
In the present work, measured position and orientation of other traffic participants are used as additional information source to improve the road course estimation. Therefore, the robust and precise tracking of other traffic participants in the vicinity of the experimental vehicle is required. For this task, a model based data fusion algorithm as presented in~\cite{GreweITS2012} is applied. Each tracked object is represented by a Kalman filter whose state vector holds the position and velocity of the traffic participant in the vehicle coordinate system. LRR, SRR and SMC sensors of the experimental vehicle provide the input for the tracking module. At each time step the tracked objects already in existence are updated with the new sensor data, while new tracked objects are created if applicable.
For the lane feature fusion only tracked objects driving in front of the experimental vehicle, that are verified by at least one radar and one camera measurement, are used to ensure high existence probabilities. Fig.~\ref{fig:lane-detection} shows tracked objects as red boxes projected onto the images of SMC and HRC.

%% file: Graphslam.tex
The aim of the lane feature fusion algorithm is to obtain a representation of the environment as a lane feature set $\mathbf{F}_{fused}$, resulting from different lane feature inputs $\mathbf{F}_1, \mathbf{F}_2, ...,\mathbf{F}_N$.
In general this problem can be described as estimation of the posterior
\begin{equation}
p(\mathbf{x}_t, \mathbf{m} | \mathbf{z}_{1:t}, \mathbf{u}_{1:t}),
\label{eq.posterior}
\end{equation}
where $\mathbf{x}_t$ is the current vehicle pose and $\mathbf{m}$ is a description of the environment
given various measurements $\mathbf{z}_{1:t}$. The control vectors $\mathbf{u}_{1:t}$ describe the movement of the vehicle at the corresponding time. This is known as Simultaneous Localization and Mapping (SLAM) problem which can be solved, for example, by an extended Kalman filter or a particle filter~\cite{thrun2002probabilistic}.

In this work, the {GraphSLAM} algorithm~\cite{Grisetti2010} is used, which estimates the environment and not only the current vehicle position $\mathbf{x}_{t}$, but the whole trajectory $\mathbf{x}_{0:t}$.
This also allows to express dependencies between measurements, which will be shown in detail later.


In GraphSLAM, eq.~\ref{eq.posterior} is described by a sparse graph.
The vehicle poses $\mathbf{x}_{0:t}$ and the environment $\mathbf{m}$ are described as vertices $\mathbf{v}_i \in \mathbf{V}$.
The measurements and control vectors describe constraints represented as edges connecting the corresponding vertices. 
The graph is formulated as the sum of constraints
\begin{equation}
J(\mathbf{V}) = \sum_{\mathbf{z}_{ij}} 
\mathbf{e}(\mathbf{z}_{ij},\mathbf{v}_i, \mathbf{v}_j)^T 
\mathbf{\Omega}_{ij} 
\mathbf{e}(\mathbf{z}_{ij},\mathbf{v}_i, \mathbf{v}_j),
\label{eq.graphslam}
\end{equation}
where 
$\mathbf{e}(\mathbf{z}_{ij},\mathbf{v}_i, \mathbf{v}_j)$
is an error function~\cite{Grisetti2010}.
This error function returns the discrepancy between the 
measurements $\mathbf{z}_{ij}$ and the vertex pose difference 
$\mathbf{\hat{z}}_{ij}(\mathbf{v}_i,\mathbf{v}_j)$.
This discrepancy is weighted by the measurement covariance in information matrix form $\mathbf{\Omega}_{ij} = \mathbf{\Sigma}_{ij}^{-1}$.
The minimization of this sum of non-linear quadratic equations can be solved by the Gauss-Newton algorithm.
The resulting configuration of vertices $\mathbf{V}^\ast = \arg \min_{\mathbf{V}} J(\mathbf{V})$ equals the poses of the estimated environment features and the vehicle poses.

%% file: Graphbuilding.tex
The environment and the measured lane features are represented as nodes in the graph $\mathcal{G}_{t}$.
Since only the environment in front of the ego vehicle is of interest, the corresponding vehicle pose set is reduced to $\mathbf{x}_{t-\tau:t}$ with $\tau + 1 $ poses.

Thus, the graph $\mathcal{G}_{t}$ contains successive vehicle poses $\mathbf{x_{t-\tau}},\mathbf{x_{t-\tau+1}},...,\mathbf{x_{t}}$
and lane features $\mathbf{f}_1, \mathbf{f}_2, ... ,\mathbf{f}_n$ as vertices $\mathbf{v}_i = [x,y,\theta]$.
Note that all poses of the graph vertices are given with respect to the current vehicle pose coordinate system.
The measurement constraints defining the edges of the graph result from the input lane features and the control vectors, which are described in the following sections.

\subsubsection{Adding odometry to the graph}
\label{odometry-graph}
The current control vector $\mathbf{u}_{t} = [\dot{\psi}, \vec{v}]^T$ is added to the previous graph $\mathcal{G}_{t-1}$.
The control vector is composed of the yawrate $\dot{\psi}$ and the speed $\vec{v}$ of the vehicle
and is used to calculate the pose difference $\mathbf{z}_{\mathbf{x}_{t-1},\mathbf{x}_{t}} = \Delta \mathbf{x}$ between the previous pose $\mathbf{x}_{t-1}$ and the current pose $\mathbf{x}_t$ with the corresponding information matrix $\mathbf{\Omega}_{\Delta \mathbf{x}}$.
At first, all vertices are transformed from $\mathbf{x}_{t-1}$ to the current vehicle pose $\mathbf{x}_{t}$ using $\Delta \mathbf{x}$.
After this transformation all vertices, arising from measurements that are more than $5~m$ behind the ego vehicle, are removed from $\mathcal{G}_{t}$.
Subsequently, past vehicle poses which are not connected to measurement vertices anymore are also removed.
The odometry edge is inserted into the graph between two successive poses as the constraint
\begin{equation}
J_t^{odo} = 
\mathbf{e}(\Delta \mathbf{x}, \mathbf{x}_{t-1}, \mathbf{x}_t)^T 
\mathbf{\Omega}_{\Delta \mathbf{x}}
\mathbf{e}(\Delta \mathbf{x}, \mathbf{x}_{t-1}, \mathbf{x}_t),
\end{equation}
with the error function 
\begin{equation*}
\mathbf{e}(\Delta \mathbf{x}, \mathbf{x}_{t-1}, \mathbf{x}_t) = [\Delta \mathbf{x} - \mathbf{\hat{z}}_{ij}(\mathbf{x}_{t-1}, \mathbf{x}_t)].
\end{equation*}
In the exemplary graph in fig.~\ref{fig:object_graph} the odometry constraint is shown as dashed black edge.

\subsubsection{Adding {SMC} clothoids to the graph}
\label{smfc-graph}
The the SMC clothoids are sampled every two meters to compute poses and information matrices of the features $\mathbf{f}^{smc}_{t,i}$ in the vehicle coordinate system.
These features are associated with all existing lane features of the graph.
If no feature within an association distance is found, a new vertex is added to the graph. 
The constraint is described as
\begin{equation}
J^{smc}_{t,i} = 
\mathbf{e}(\mathbf{f}^{smc}_{t,i}, \mathbf{x}_{t}, \mathbf{f}^{\ast})^T 
\mathbf{\Omega}^{smc}
\mathbf{e}(\mathbf{f}^{smc}_{t,i}, \mathbf{x}_{t}, \mathbf{f}^{\ast}),
\end{equation}
where the measurement $\mathbf{f}^{smc}_{t,i}$ is the desired pose difference
between the vertex of the current vehicle pose $\mathbf{x}_{t}$ and the vertex of the new or associated feature $\mathbf{f}^{\ast}$.
The measurement of an SMC feature associated to an existing feature is shown in fig.~\ref{fig:object_graph} as a green edge.

\subsubsection{Adding {HRC} features to the graph}
\label{hrc-graph}
Since features in the HRC images are extracted directly at lane markings,
the corresponding features $\mathbf{f}^{hrc}_{t,i}$ in vehicle coordinates are directly associated with an existing feature or inserted as a new vertex with the corresponding measurement constraint
\begin{equation}
J_{t,i}^{hrc} = 
\mathbf{e}(\mathbf{f}^{hrc}_{t,i}, \mathbf{x}_{t}, \mathbf{f}^{\ast})^T 
\mathbf{\Omega}^{hrc}
\mathbf{e}(\mathbf{f}^{hrc}_{t,i}, \mathbf{x}_{t}, \mathbf{f}^{\ast}),
\end{equation}
which is shown in fig.~\ref{fig:object_graph} as magenta colored edge.
\subsubsection{Adding dynamic object features to the graph}
\label{dynamic-objects-graph}
The idea is to use the positions and movement of other traffic participants to derive information about the lanes.
In most cases drivers of other vehicles tend to drive near the middle of the lane.
Based on this assumption, lane features are generated from tracked dynamic objects.
Two features perpendicular to the object heading are generated on the left and right side of each dynamic object with a distance of $w/2$ representing potential lane markings.
The parameter $w$ is an estimation of the current lane width that is taken from the $SMC$ clothoids if available or assumed to be $3.5~m$ \footnote{Most common lane width on German highways.} otherwise.

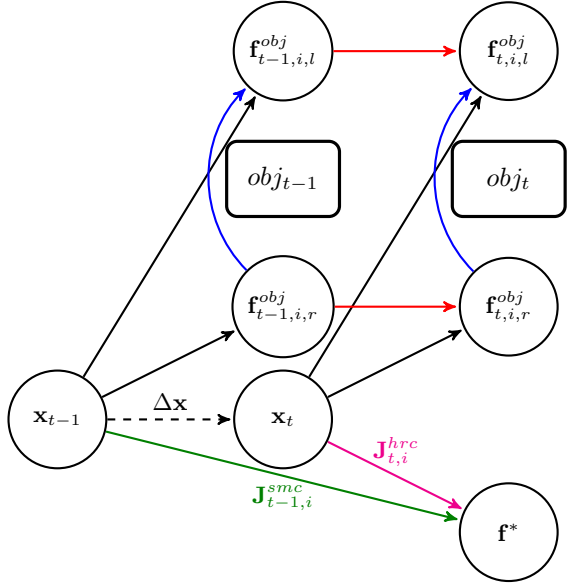
\begin{figure}[t]
	\begin{tikzpicture}[->,>=stealth',shorten >=1pt,auto,node distance=3cm,
	thick,main node/.style={circle,draw,minimum size=1.3cm,font=\sffamily\small}]
	
	\node[main node] (1) {$\mathbf{x}_{t-1}$};
	\node[main node] (2) [right of=1] {$\mathbf{x}_{t}$};

	\node[box] (3) [above of=2, node distance=3.2cm] {$obj_{t-1}$};
	
	\node[main node] (4) [above of=3,node distance=1.7cm] {$\mathbf{f}^{obj}_{t-1,i,l}$};
	\node[main node] (5) [below of=3,node distance=1.7cm] {$\mathbf{f}^{obj}_{t-1,i,r}$};
	
	\node[box] (6) [right of=3, node distance=3cm] {$obj_{t}$};
	
	\node[main node] (7) [above of=6,node distance=1.7cm] {$\mathbf{f}^{obj}_{t,i,l}$};
	\node[main node] (8) [below of=6,node distance=1.7cm] {$\mathbf{f}^{obj}_{t,i,r}$};
	
	\node[main node] (9) [below of=8,node distance=3cm] {$\mathbf{f}^{\ast}$};
	
	\path[every node/.style={font=\sffamily\small}]
	(1) edge[black, dashed] node [above] {$\Delta \mathbf{x}$} (2)
	(1) edge[black] node [above] {} (4)
	(1) edge[black] node [above] {} (5)
	(2) edge[black] node [above] {} (7)
	(2) edge[black] node [above] {} (8)
	(4) edge[red, ] node [above] {} (7)
	(5) edge[red, ] node [above] {} (8)
	(5) edge[blue,  bend left = 45] node [above] {} (4)
	(8) edge[blue,  bend left = 45] node [above] {} (7)
	(1) edge[black!50!green ] node [below] {$\mathbf{J}^{smc}_{t-1,i}$} (9)
	(2) edge[magenta] node [above] {$\mathbf{J}^{hrc}_{t,i}$} (9)
	;

	
	\end{tikzpicture}
	\centering
	\caption[A GraphSLAM example]{Example graph of measuring an object over two successive time steps $t-1$ and $t$ and the feature $\mathbf{f}^{\ast}$ obtained from the SMC and the HRC with the odometry edge (dashed black), left and right object feature edges (black) width edges (blue), smoothing edges (red), the SMC measurement edge (green) and the HRC measurement edge (magenta).}
	\label{fig:object_graph}
\end{figure}

The corresponding feature covariance equals the sum of the object covariance and 
a covariance matrix representing the lateral standard deviation of traffic participants with respect to the middle of a lane. 
The resulting features are associated with existing features or added as a new vertex with the measurement constraint
\begin{equation}
J_{t,i,l/r}^{obj} = 
\mathbf{e}(\mathbf{f}^{obj}_{t,i,l/r}, \mathbf{x}_{t}, \mathbf{f}^{\ast})^T 
\mathbf{\Omega}^{obj}
\mathbf{e}(\mathbf{f}^{obj}_{t,i,l/r}, \mathbf{x}_{t}, \mathbf{f}^{\ast}),
\end{equation}
where $\mathbf{f}^{obj}_{t,i,l/r}$ is the left or right feature of the $i$-th tracked object at timestamp $t$.
An example of measurements of one tracked object over two successive time steps is shown in fig.~\ref{fig:object_graph}. 

A deficiency of this model is that left and right features are decoupled,
which means that an improvement of the position of the left feature does not influence the right one and vice versa. 
Therefore, the assumption on the lane width is expressed as a constraint between the left and right feature:
\begin{equation}
J_{t,i}^{width} = 
\mathbf{e}(\mathbf{w}, \mathbf{f}^{obj}_{t,i,l}, \mathbf{f}^{obj}_{t,i,r})^T 
\mathbf{\Omega}^{width}
\mathbf{e}(\mathbf{w}, \mathbf{f}^{obj}_{t,i,l}, \mathbf{f}^{obj}_{t,i,r}).
\end{equation}
The desired pose difference between the left and right feature of the same object is defined as 
$\mathbf{w} = [0,w,0^{o}]$ with the lane width $w$ as lateral distance.
The angle difference is set to zero, since the heading of the features is supposed to be equal.
The information matrix $\mathbf{\Omega}_{width}$ corresponds to the variance of the current lane width estimation. 
The corresponding constraints are shown in fig.~\ref{fig:object_graph} as blue edges.

Furthermore, one additional dependency has to be considered: in the current model,
two successive features on the same side of a tracked object are decoupled,
this means $\mathbf{f}^{obj}_{t-1,i,l}$ has no direct influence on $\mathbf{f}^{obj}_{t,i,l}$. 
If a feature $\mathbf{f}^{obj}_{t-1,l}$ is corrected by other measurements,  a large discrepancy to the succeeding feature can occur, which needs to be minimized.
Therefore, a smoothing constraint
\begin{equation}
J_{t,i,l}^{smo} = 
\mathbf{e}(\mathbf{0},\mathbf{f}^{obj}_{t-1,l},\mathbf{f}^{obj}_{t,l})^T 
\mathbf{\Omega}^{smo}
\mathbf{e}(\mathbf{0},\mathbf{f}^{obj}_{t-1,l},\mathbf{f}^{obj}_{t,l})
\label{eq.smoothing}
\end{equation}
is added between the two features.
The lateral displacement between successive features can then be reduced by increasing  $\mathbf{\Omega}_{yy}^{smo}$ and $\mathbf{\Omega}_{\theta \theta}^{smo}$.
Note that $\mathbf{\Omega}_{xx}^{smo} = 0$, since the longitudinal distance is not supposed to be altered.  
In fig.~\ref{fig:object_graph} this constraint is shown as red edge.
If traffic participants perform a lane change this constraint is strongly violated, since at some point the feature of $\mathbf{f}^{obj}_{t-1,l}$ belongs to one lane and the features of $\mathbf{f}^{obj}_{t,l}$ to the neighboring lane. Here GraphSLAM provides the utility to multiply eq.~\ref{eq.smoothing} with a switch variable 
$0 \leq s_{t,i,l/r}  \leq 1$. 
If this variable is set to zero, the edge is disabled and if it equals one, it is fully activated.
As in~\cite{sunderhauf2012switchable} this method is used for false loop closures during the optimization of GraphSLAM, where 
$J_{t,i,l/r}^s = \Omega^s(1-s_{t,i,l/r})^2$ is added as a further constraint.
This forces the edge to be enabled until the error of the edge gets too large and deactivating the edge will become more optimal. 
\subsubsection{Solving the graph}
In summary, the graph $\mathcal{G}_{t}$ consists of the constraints
\[ 
J(\mathbf{v}) = \sum_{t} J_t^{odo} + \sum_t\sum_i J_{t,i}^{smc} + \sum_t\sum_i J_{t,i}^{hdc}  
\]
\begin{equation}
+ \sum_t \sum_i (J_{t,i,l/r}^{obj} + J_{t,i}^{width} +  s_{t,i,l/r} J_{t,i,l/r}^{smo} + J_{t,i,l/r}^s),
\end{equation}
where $t$ sums over all $\tau+1$ relevant time steps and $i$ over all sensor features at the corresponding time step.

\begin{figure}[t]
	\begin{center}
		\includegraphics[width=8.5cm]{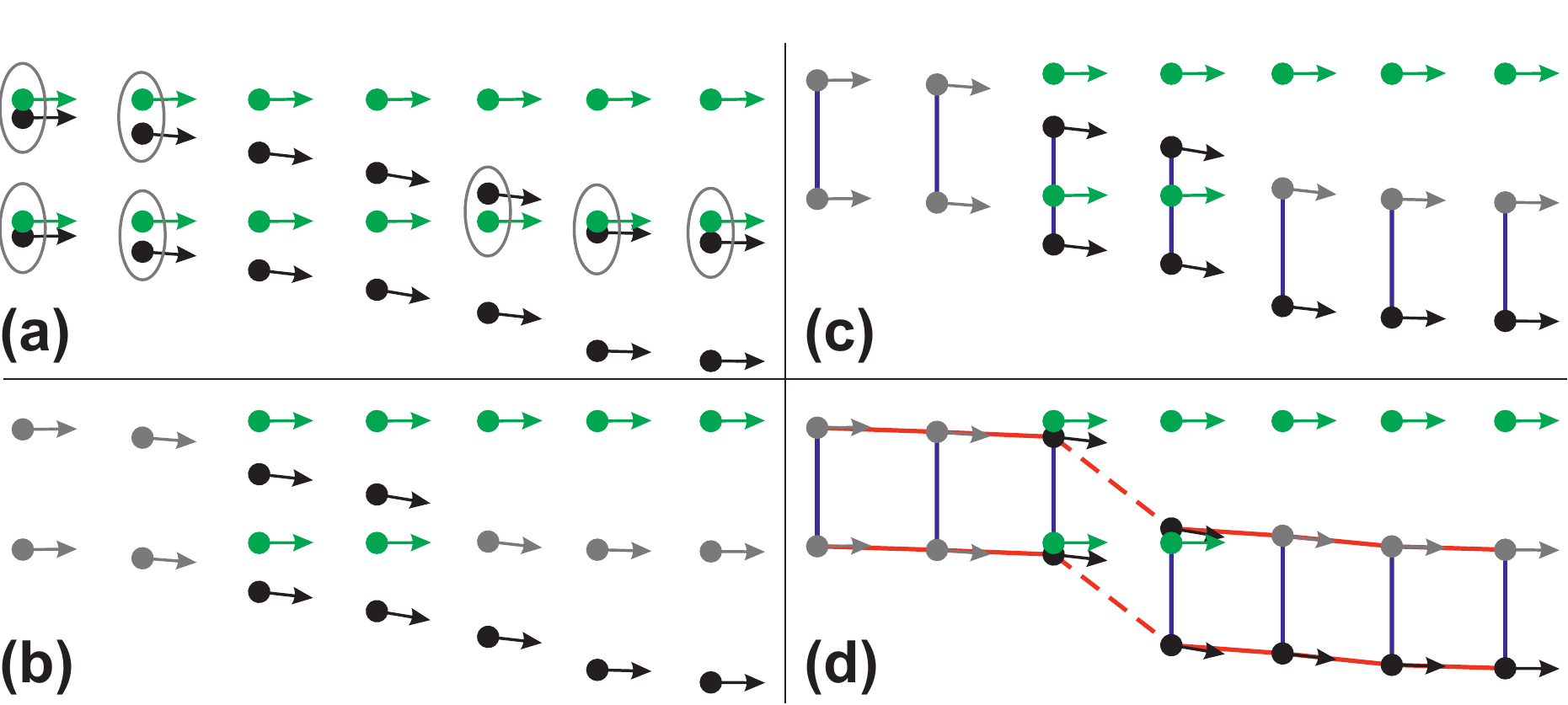}
		\caption{Example of the fusion of SMC and HRC lane features (green) with the features generated on the left and right side of a tracked object (black), which performs a lane change to the right neighboring lane.
		(a) Features are associated during the graph building (gray ellipse).
		(b) Resulting features after solving GraphSLAM without any lane width $J^{width}$ and smoothing $J^{smo}$ edges and the fused features (gray). 
		(c) Solution with lane width edges (blue) between object features. Here the feature distances still equal the lane width after solving the graph.
		(d) Result from solving the graph with smoothing edges (red). During the optimization two smoothing edges got disabled (dashed red) by the switch variables.
					\label{fig:graphslamfusion}}
	\end{center}
\end{figure}

A configuration of optimal vertex poses is obtained by solving the graph. The result of the algorithm is represented as a set of fused lane features $\mathbf{F}_{fused}$ which correspond directly to these optimal vertex poses. Note that the confidence values $c$ of the resulting fused features are updated whenever measurement features are associated. The influence of the smoothing and width constraints is illustrated in fig.~\ref{fig:graphslamfusion}.
Note that an ego lane change has no impact on these constraints, since it only influences the odometry edges.

\begin{figure*}
	\begin{center}
		\includegraphics[width=17.0cm]{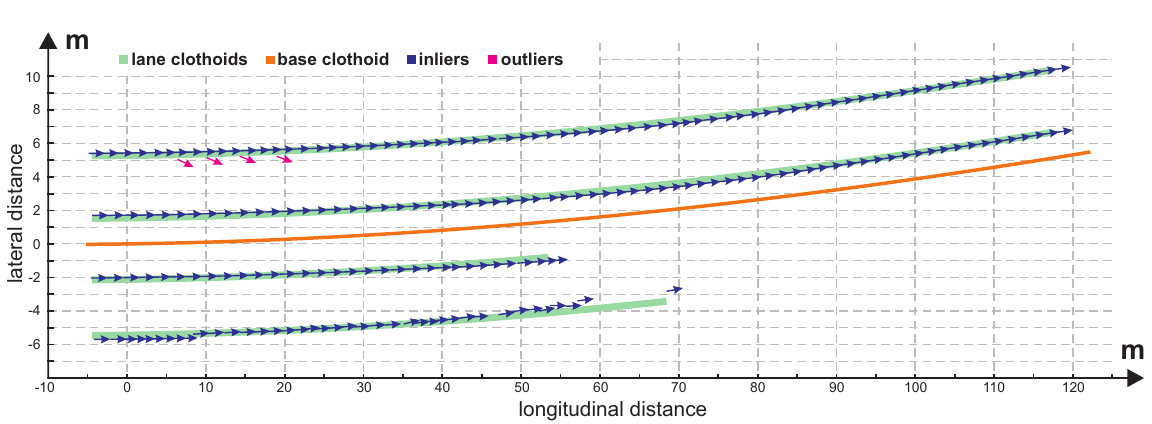}
		\caption{Estimation of the base clothoid (orange) representing the road course and the lanes modeling (light green) based on the fused features. Valid features of the base clothoid fit are shown in blue while outliers are depicted in magenta. In this snap-shot features are generated from three tracked objects at approximately $(x,y) = [(120,8.5), (55,1), (9,-4)]$. Features of the adjacent lane to the right (negative $y$-direction) with a distance larger than $9~m$ are provided solely from the HRC. \label{fig:clothoid-fitting}}
	\end{center}
\end{figure*}

%% file: Roadmodel.tex
The resulting fused lane features are used to extract the parameters of a mathematical model to represent ego lane and neighboring lanes. Such a model should be able to handle the curved roads and is supposed to be robust against measurement noise, missing data and outliers. Various models have been proposed to model the road and traffic lanes: straight lines~\cite{Lakshmanan1996}, parabolic curves~\cite{Liu2011}, Euler spirals also called clothoids~\cite{Dickmanns1992,Meis2010} and splines~\cite{Wang2000,Aly2008,Kim2008}.

Although simple models, such as straight lines, are very robust against noise, they cannot model real roads with desired distance and accuracy. On the contrary, more complex techniques, such as parabolic curves, clothoids or splines, can accurately model complicated road shapes, but these models are also more sensitive to uncertainties in the measured data and require sophisticated methods for noise reduction and filtering.

In the present work, the clothoid model is used for modeling course of the road and traffic lanes, since this model is used in the design and construction of highways. Clothoids are defined by a linear change of curvature over the arc length $x$. For small heading angles (up to $15$\textdegree) clothoids can be sufficiently precise approximated by a third order polynomial~\cite{Dickmanns1992}:
\begin{equation}
y(x) = y_{0} + \theta_{0}x + c_{0}x^{2}/2 + c_{1}x^{3}/6
\label{eq:clothoid-model},
\end{equation}
where $y_{0}$ is the lateral offset of the clothoid, $\theta_{0}$ and $c_{0}$ are heading and curvature at $x = 0$ and $c_{1}$ describes the change of curvature.

The first step towards modeling traffic lanes is the estimation of the road course. For this purpose a base clothoid is computed by fitting the heading of the fused lane features using the derivative of eq.~\ref{eq:clothoid-model}.
The parameters $\theta^{base}_{0}$, $c^{base}_{0}$ and $c^{base}_{1}$ of the base clothoid are estimated using Robust Linear Least Squares regression~\cite{Lawson1987}. This is an iterative approach which detects and eliminates outliers from the fitting procedure. The confidence values $c$ of the features serve as weights in the fit. Fig.~\ref{fig:clothoid-fitting} shows the estimated base clothoid (orange) for a set of input features. The features considered for modeling are depicted in dark blue, while features classified as outliers are drawn in magenta.

\begin{figure}[t]
	\begin{center}
		\includegraphics[width=8.6cm]{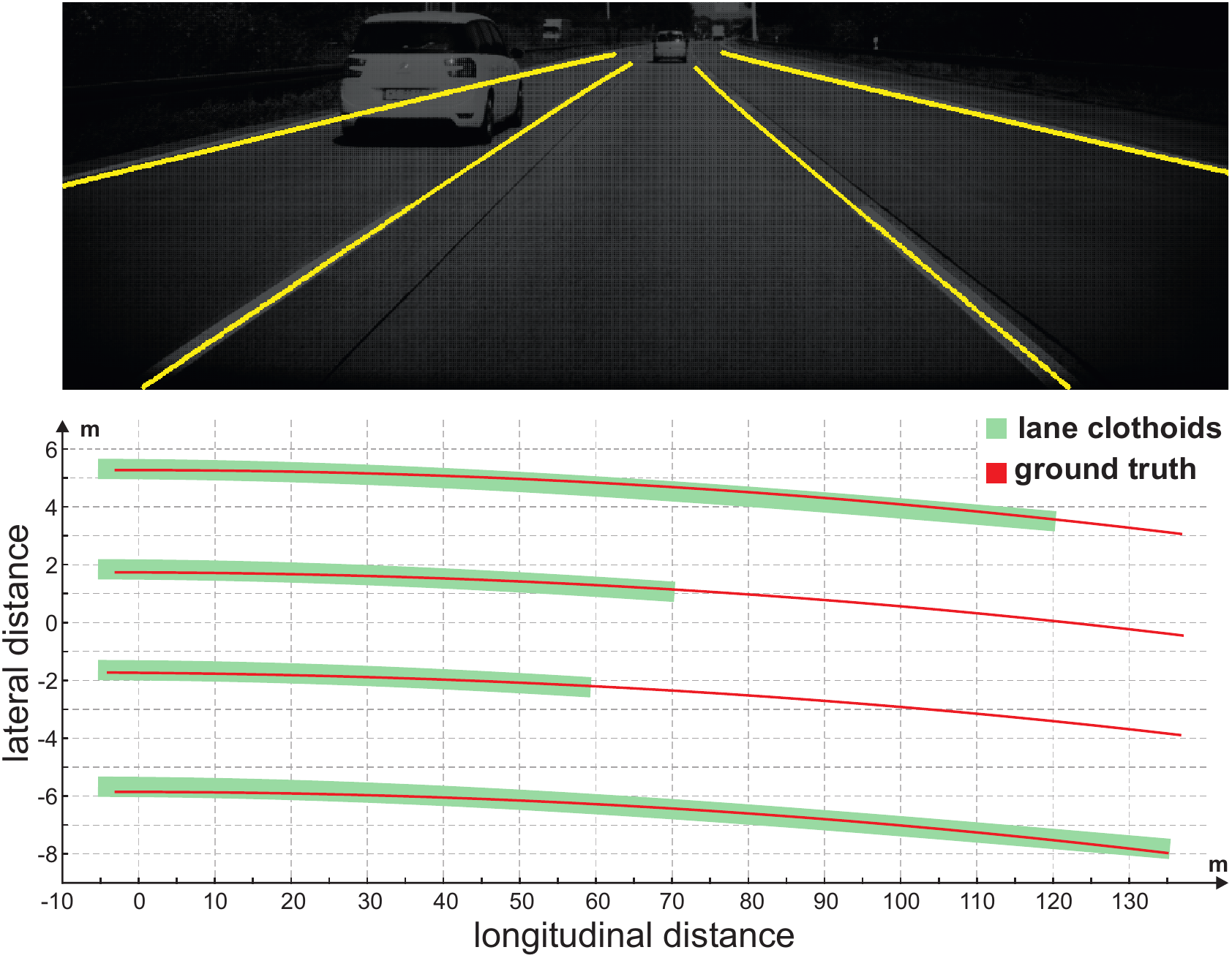}
		\caption{Exemplary result of the modeled lane clothoids projected in yellow onto the camera image (top). The same result is shown in green in the top view diagram (bottom). The red curves represent the ground truth data needed for comparison and evaluation of the result. \label{fig:final-lanes}}
	\end{center}
\end{figure}

\begin{figure}[t]
	\begin{center}
		\includegraphics[width=8.6cm]{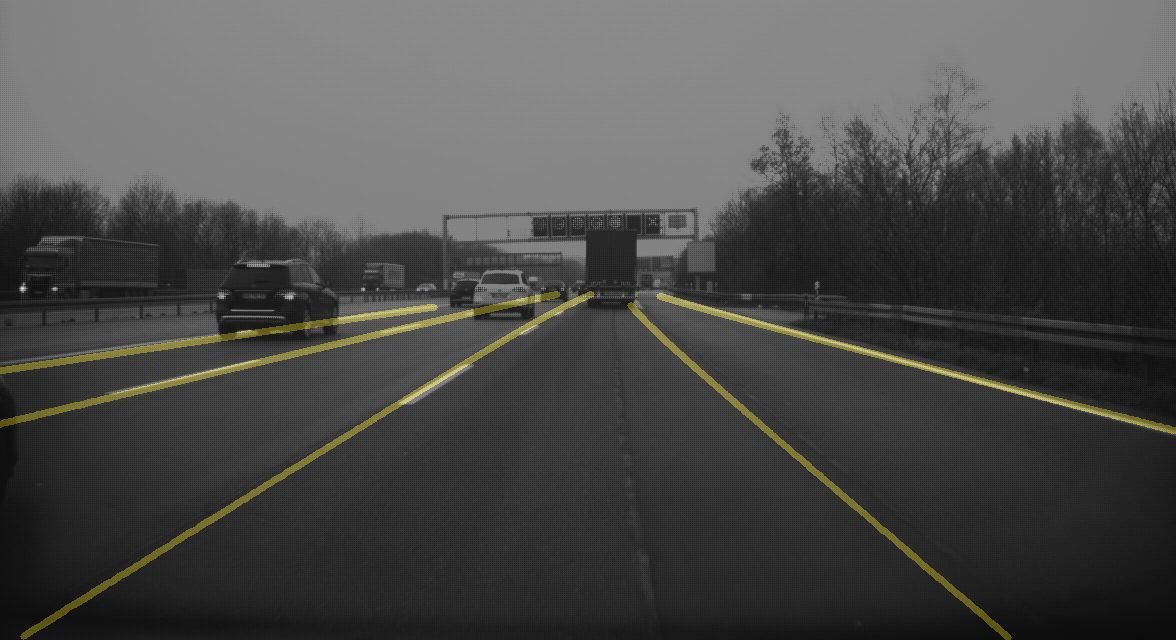}
		\caption{Estimated lane clothoids (yellow) projected onto the camera image in a scenario with three drivable lanes, an emergency lane and several vehicles. Despite missing markers on the right side of the currently driven lane the presented method is able to obtain a stable result. Note that in this scene a successful estimation of the current lane is only accomplished due to the usage of the tracked trajectory of the truck driving in front of the ego vehicle.\label{fig:missing-marker}}
	\end{center}
\end{figure}

Once the course of the road is known, the offsets of the lane clothoids need to be determined. To obtain the offset of a lane clothoid, it is necessary to know which of the fused lane features belong to that lane. If there already was an estimated lane model at the previous time step, that information is used to group the features. New clothoids are created by combining the offsets of clothoids from the previous time step with the parameters of the current base clothoid:
\begin{equation}
 y(x) = y^{t-1}_{0} + \theta^{base}_{0}x + c^{base}_{0}x^{2}/2 + c^{base}_{1}x^{3}/6
 \label{eq:cluster-cloth}.
\end{equation}
Each feature is associated to the closest of these clothoids resulting in several feature groups. Usually some features (all features in the first algorithm loop) cannot be associated using this clustering attempt, since no corresponding clothoid from the previous time step exists. Therefore, remaining features in the near range of the experimental vehicle are projected onto the y-axis. If enough features' projections are in close proximity to each other, the mean lateral position is computed and used as offset in eq.~\ref{eq:cluster-cloth} instead of $y^{t-1}_{0}$ to group the remaining features. In the last step, a Least Squares fit is performed for each group of features, that yields the offset of a lane clothoid. Thus the lane clothoids of the current time step are composed of the fitted offsets in conjunction with the course parameters of the base clothoid. To obtain a stable and continuous description of the lanes, a Kalman filter is used to filter each lane clothoid over time. An example of the resulting lane clothoids projected onto the SMC image is illustrated in yellow in the upper part of fig.~\ref{fig:final-lanes}. The lower part shows a top view diagram of the same clothoids plotted in green. Fig.~\ref{fig:missing-marker} shows a scenario where lane markers are missing on the right side of th ecurrent lane. The lack of visual information is compensated by the utilization of trajectories of other traffic participants resulting in a stable lane estimation.\footnote{A video sequence illustrating the results is available as supplementary material.}

%% file: Evaluatiomethod.tex
\begin{figure}[h]
	\begin{center}
		\includegraphics[width=8.6cm]{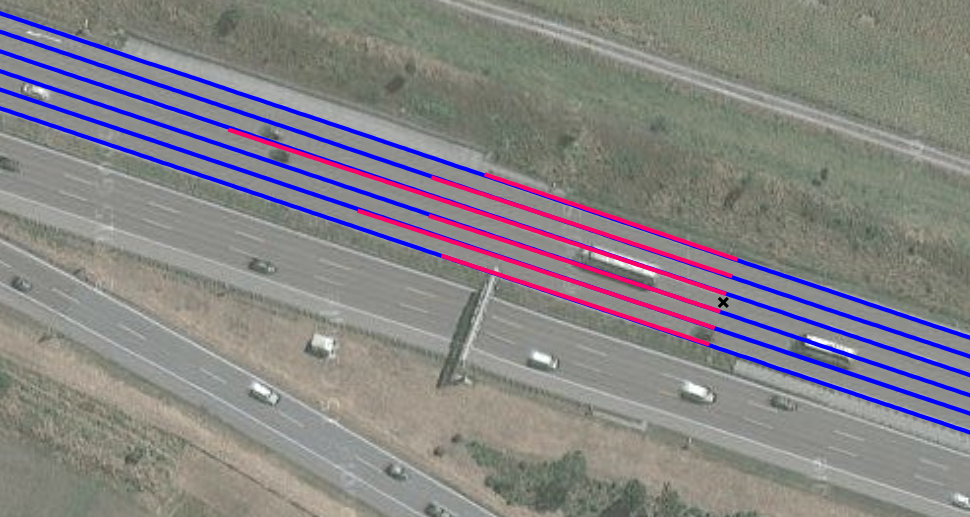}
		\caption[google maps plot]{Aerial view of a short extract of Germany's A9 highway~\protect\footnotemark taken from Google maps. The modeled lane clothoids (red) and ground truth data (blue) are plotted onto the map and the current position of the experimental vehicle is denoted by the black cross. Note that the vehicles visible in the map are not correlated to the traffic participants used in the estimation of the lane clothoids.}
		\label{fig:lanes-google-map}
	\end{center}
\end{figure}

The performance of the lane feature fusion and subsequent lane clothoid modeling are evaluated using real data collected with the experimental vehicle on highways. For comparison a ground truth map containing global positions of lane markings of a highway route with multiple traffic lanes and left and right curves is used. The ground truth map is generated by driving on the highway on all traffic lanes using a RT4000 GPS system together with the odometry of the experimental vehicle and lane markings detected by the serial camera. The RT4000 family products are advanced, precision inertial and GPS Navigation systems for measuring motion, position and orientation~\footnote{Oxford {T}echnical {S}olutions, {R}{T}4000 family systems, \url{http://www.oxts.com/products/rt4000-family/}}. The separate data sets collected from each lanes are merged offline and the RT4000 GPS system is used again for localization in the ground truth map. A snap-shot of the ground truth map (red) and the modeled lane clothoids (green) is shown in the lower part of fig.~\ref{fig:final-lanes}. Another snap-shot of estimated lane clothoids and ground truth data plotted onto a map extract is shown in fig.~\ref{fig:lanes-google-map}. To compare the estimated lanes to the ground truth map, clothoids are sampled every $10~m$ and the lateral deviation at each sampled distance to the ground truth map is computed. This is done after every algorithm loop resulting in a distribution of measurements at each sample distance. For the evaluation the data collected during several drives on the highway with average traffic density is analyzed corresponding to a total driving distance of $24~km$. As the measure of performance the mean, standard deviation and Root Mean Square Error (RMSE) of the distributions at each sampled distance accumulated over the total data are computed. The evaluation is conducted up to a maximum distance of $120~m$ separately for ego and adjacent lanes as they are expected to show different results.

\begin{table*}
	\begin{center}
		\caption{Mean, sigma and RMSE for ego and adjacent lane.}
		\begin{tabular}{ c |  c  c  c |  c  c  c }
			\midrule
			distance & \multicolumn{3}{c |}{ego lane} & \multicolumn{3}{c }{adjacent lane} \\
			~[m]~ & $\mu$ [m] & $\sigma$ [m] & $RMSE$ [m] & $\mu$ [m] & $\sigma$ [m]  & $RMSE$ [m]\\
			\midrule
			0 &  0.00 & 0.10 & 0.10 & -0.04 & 0.20 & 0.21 \\
			20 & -0.02 & 0.10 & 0.11 & -0.07 & 0.20 & 0.21 \\
			40 & -0.06 & 0.17 & 0.18 & -0.11 & 0.25 & 0.27 \\
			60 & -0.11 & 0.26 & 0.28 & -0.17 & 0.32 & 0.37 \\
			80 & -0.20 & 0.38 & 0.42 & -0.24 & 0.44 & 0.50 \\
			100 & -0.26 & 0.48 & 0.55 & -0.29 & 0.59 & 0.66 \\
			120 & -0.26 & 0.58 & 0.64 & -0.47 & 0.87 & 0.99 \\
			\midrule
		\end{tabular}
		\label{tab:result}
	\end{center}
\end{table*}

\begin{figure}[t]
	\begin{center}
		\includegraphics[width=8.7cm]{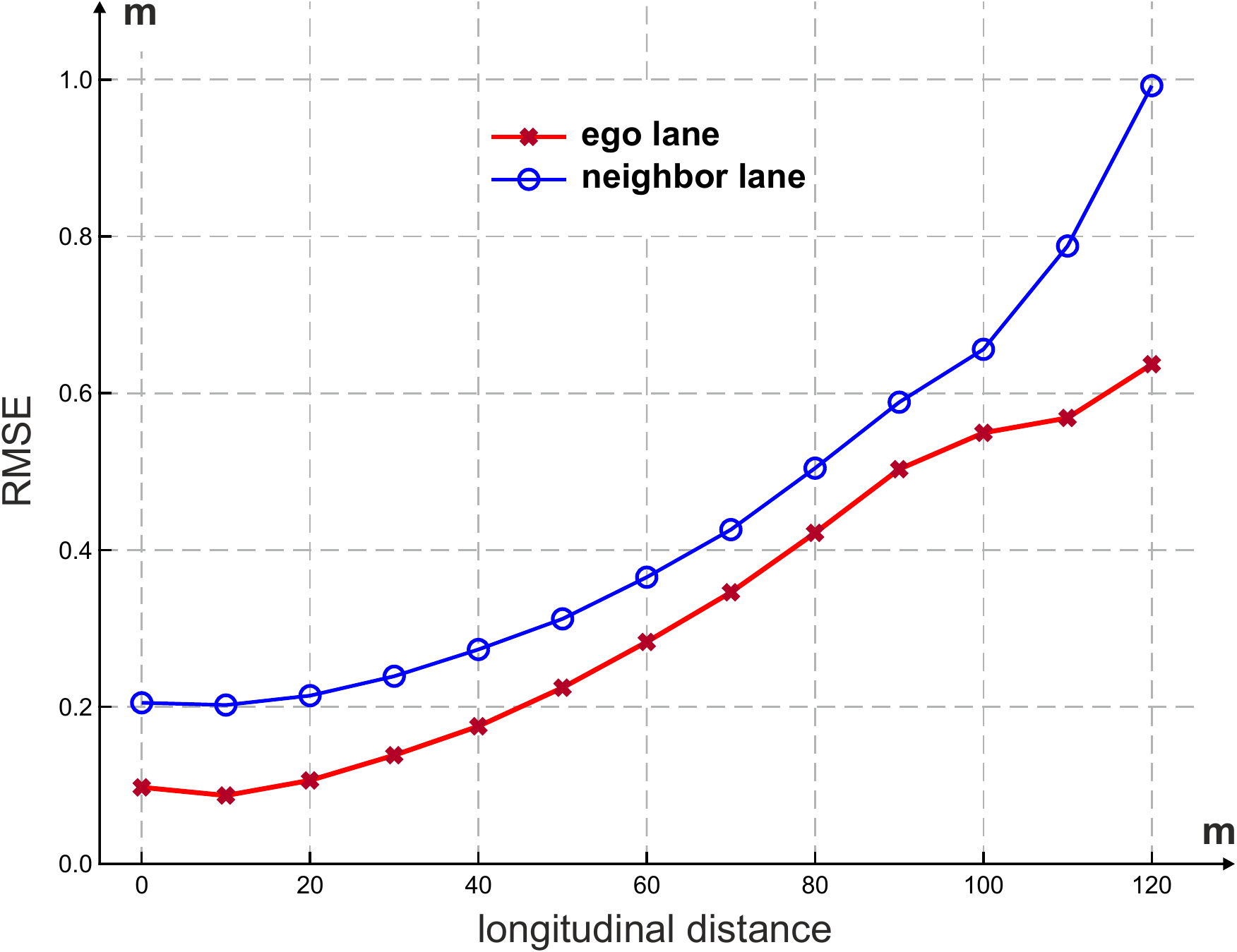}
		\caption{RMSE of the difference between estimated lanes and ground truth for ego (red) and neighbor lane (blue).}
		\label{fig:rmse}
	\end{center}
\end{figure}

Results of the evaluation are presented in table~\ref{tab:result} and the RMSE for ego (red) and adjacent lanes (blue) is plotted in fig.~\ref{fig:rmse}. While mean lateral deviation between estimated lanes and ground truth data is below one sigma for all analyzed distances, one can see that their quality decreases with distance. Due to data fusion over several time steps, lane features near the ego vehicle have been measured more often and have therefore higher accuracy. The results also show that the ego lane is estimated with a better performance than the adjacent lanes. One reason for this is that the SMC only contributes features to the lane currently driven on. At larger distances the estimation of the ego lane is at all times dominated by the tracked object right in front of the experimental vehicle and HRC data up to that object. The adjacent lanes on the other hand are often estimated using either tracked objects or HRC data depending on the current traffic situation.

In summary the lane feature fusion and the subsequent road modeling provide estimation of lanes on highways in real-time (less than $100 ~ms$). Using the SMC, the HRC and tracked objects as input, an increase in the detection range of the ego lane compared to the serial lane detection of the SMC (up to $90 ~m$) is achieved. At $120 ~m$ the ego lane is estimated with an RMSE of $0.64 ~m$. Due to the combination of HRC data and tracked objects neighboring lanes are estimated at all times. At a distance of $120 ~m$ the RMSE obtained for neighboring lanes is less than $1 ~m$.

It needs to be noted that evaluation on publicly available datasets such as \textit{KITTI}~\cite{Fritsch2013ITSC}, \textit{DARPA Urban Challenge}~\cite{Huang2010IJRR}, \textit{Caltech}~\cite{CalTechOnline}, \textit{ROMA}~\cite{Veit2008} is not reasonable for the presented method. None of the datasets contains all of the following items that would be necessary for a meaningful comparison: camera images, ground truth data of multi-lane roads, tracked trajectories of other traffic participants and odometry of the ego vehicle.

\footnotetext{The A9 autobahn in Germany is declared  to be a test track for self-driving cars by Germany's ministry of transport~\cite{GermanMinistryOnline}.}

%% file: Conclusion.tex
In the present work, a real-time perception method for highways estimating ego and neighbor lanes is presented. In addition to the serial camera system of the vehicle, a high resolution camera system and information deduced from the tracking of other traffic participants serve as input to the GraphSLAM based feature fusion algorithm. The fusion result is utilized to model traffic lanes on a highways represented as clothoids. No prior knowledge about the road or lanes is needed in the feature fusion and lane modeling procedure. Evaluation of the approach is conducted by comparing the estimated lanes to a ground truth map of a several kilometer long route on a highway. In comparison to the serial lane detection, the method provides increased detection range for the ego lane and detection of the neighbor lanes at all times. 

The presented method works especially well in scenarios where many traffic participants are present. Hidden lane markings are compensated by using tracked trajectories of other vehicles in combination with camera data.

The results show that up to a distance of $120 ~m$ ahead of the vehicle good estimates of ego and adjacent lanes are obtained. Precision, stability and robustness of the derived multi-lane model suffice to be used in the development and testing of self-driving cars on highways. Up to now, several hundred kilometers have been driven autonomously on highways applying the presented methods in the experimental vehicle.

Enhancements to the presented lane perception method could be achieved by increasing the quality of the input data, e.g. improving outlier suppression in the lane feature extraction from the HRC images at large distance. Furthermore, the addition of new data sources such as vehicle-to-vehicle communication could lead to an increase in performance and robustness.
The proposed method for multi-lane modeling yields a continuous and robust lane description on highways by initially estimating a base clothoid, that characterizes the general road course. However, a drawback of this approach is that highway exit and entrance ramps as well as highway junctions can not be represented in full detail. Development of a spline based lane model could pose a solution to this issue, as splines provide a more flexible description of the lanes and might therefore be able to cover more complex scenarios.